\documentclass[10pt,twocolumn,letterpaper]{article}
\usepackage{lmodern}
\usepackage{siunitx}
\usepackage{booktabs}
\usepackage{etoolbox}

\usepackage{cvpr}
\usepackage{times}
\usepackage{epsfig}
\usepackage{graphicx}
\usepackage{amsmath}
\usepackage{amssymb}

\newcommand*{\affaddr}[1]{#1} 
\newcommand*{\affmark}[1][*]{\textsuperscript{#1}}
\newcommand*{\email}[1]{\texttt{#1}}

\usepackage{multirow}

\usepackage{array}
\newcolumntype{L}[1]{>{\raggedright\let\newline\\\arraybackslash}m{#1}}
\newcolumntype{C}[1]{>{\centering\let\newline\\\arraybackslash}m{#1}}
\newcolumntype{R}[1]{>{\raggedleft\let\newline\\\arraybackslash}m{#1}}
\usepackage{pifont}
\usepackage{amsmath}
\usepackage{algorithm}
\usepackage[noend]{algpseudocode}
\usepackage{float}
\usepackage{placeins}
\newcommand{\tablestyle}[2]{\setlength{\tabcolsep}{#1}\renewcommand{\arraystretch}{#2}\centering\footnotesize}

\newcommand{\bd}[1]{\textbf{#1}}
\newcommand{\app}{\raise.17ex\hbox{$\scriptstyle\sim$}}

\newcolumntype{x}[1]{>{\centering\arraybackslash}p{#1pt}}

\newlength\savewidth\newcommand\shline{\noalign{\global\savewidth\arrayrulewidth
  \global\arrayrulewidth 1pt}\hline\noalign{\global\arrayrulewidth\savewidth}}
\makeatletter\renewcommand\paragraph{\@startsection{paragraph}{4}{\z@}
  {.5em \@plus1ex \@minus.2ex}{-.5em}{\normalfont\normalsize\bfseries}}\makeatother

\usepackage{times,graphicx,amsmath,amssymb,booktabs,tabulary,multirow,overpic,xcolor}
\usepackage[caption=false]{subfig}
\usepackage[british,english,american]{babel}

\usepackage[pagebackref=true,breaklinks=true,letterpaper=true,colorlinks,bookmarks=false]{hyperref}

\cvprfinalcopy 

\def\cvprPaperID{****} 

\ifcvprfinal\fi
\begin{document}

\title{Towards Universal Object Detection by Domain Attention}

\author{%
Xudong Wang\affmark[1], Zhaowei Cai\affmark[1], Dashan Gao\affmark[2] and Nuno Vasconcelos\affmark[1]\\
\affaddr{\affmark[1]University of California, San Diego}, \affaddr{\affmark[2]12 Sigma Technologies}\\
\email{\{xuw080,zwcai,nuno\}@ucsd.edu, dgao@12sigma.ai}
}
\maketitle

\begin{abstract}
Despite increasing efforts on universal representations for visual recognition, few have addressed object detection. In this paper, we develop an effective and efficient universal object detection system that is capable of working on various image domains, from human faces and traffic signs to medical CT images. Unlike multi-domain models, this universal model does not require prior knowledge of the domain of interest. This is achieved by the introduction of a new family of adaptation layers, based on the principles of squeeze and excitation, and a new domain-attention mechanism. In the proposed universal detector, all parameters and computations are shared across domains, and a single network processes all domains all the time. Experiments, on a newly established universal object detection benchmark of 11 diverse datasets, show that the proposed detector outperforms a bank of individual detectors, a multi-domain detector, and a baseline universal detector, with a 1.3$\times$ parameter increase over a single-domain baseline detector. The code and benchmark will be released at http://www.svcl.ucsd.edu/projects/universal-detection/.
\end{abstract}
\section{Introduction}
\label{sec:intro}

\begin{figure}
  \includegraphics[width=\linewidth]{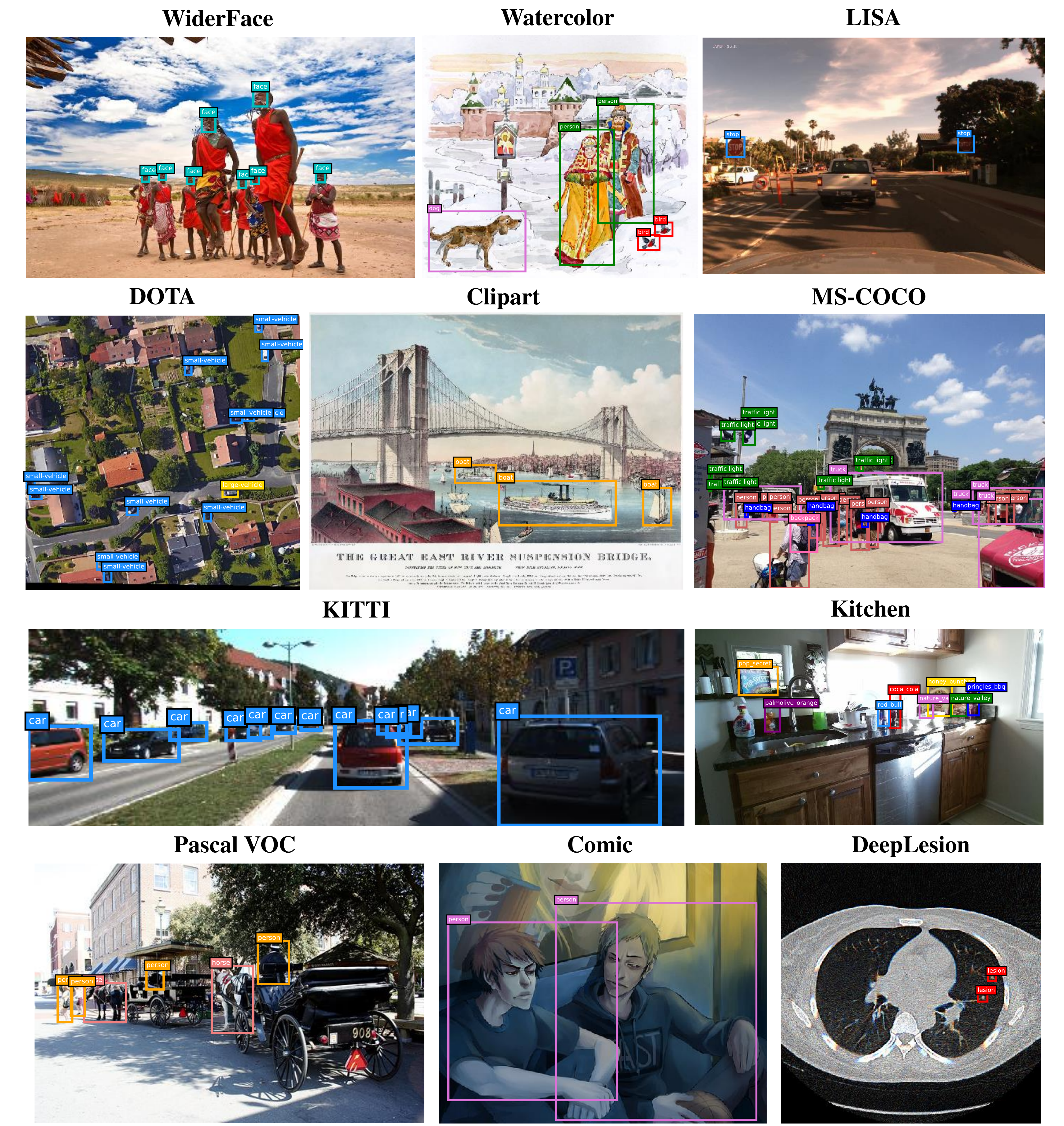}
  \caption{Samples of our universal object detection benchmark.}
\label{fig:demo}\vspace{-5mm}
\end{figure}

\begin{figure*}[!t]
\begin{minipage}[t]{.402\linewidth}
\small
\centering
\centerline{\includegraphics[width=0.95\linewidth]{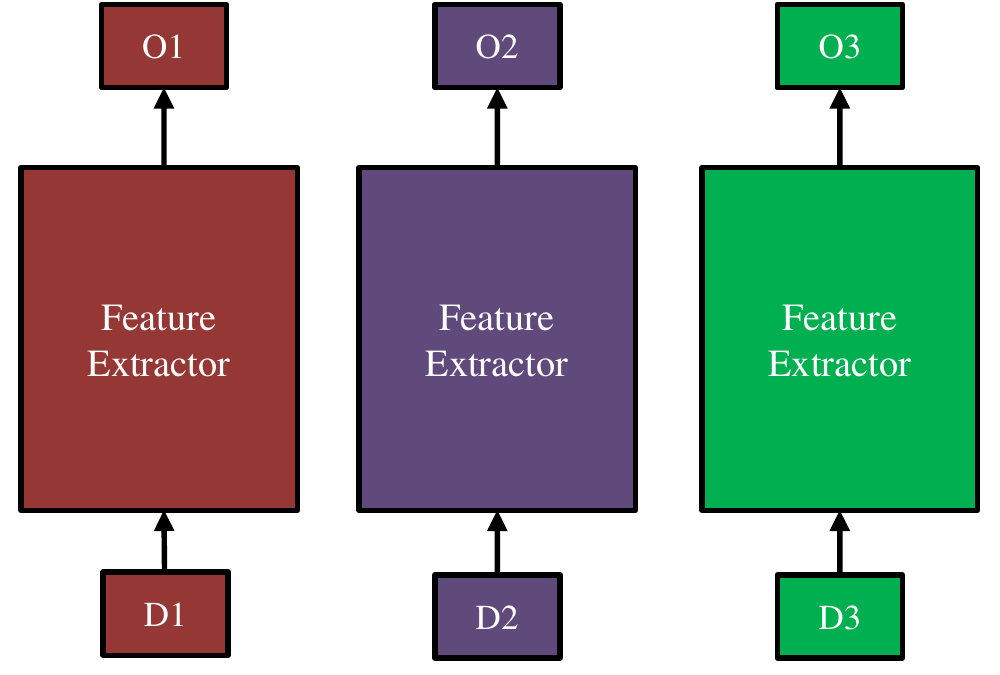}}{(a) Single-domain Detector Bank}
\end{minipage}
\hfill
\begin{minipage}[t]{.18\linewidth}
\small
\centering
\centerline{\includegraphics[width=0.95\linewidth]{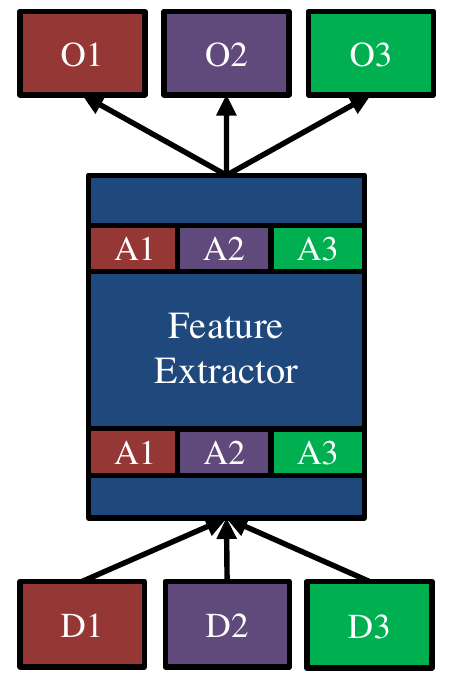}}{(b) Adaptive Multi-domain Detector}
\end{minipage}
\hfill\vline\hfill
\begin{minipage}[t]{.18\linewidth}
\small
\centering
\centerline{\includegraphics[width=0.95\linewidth]{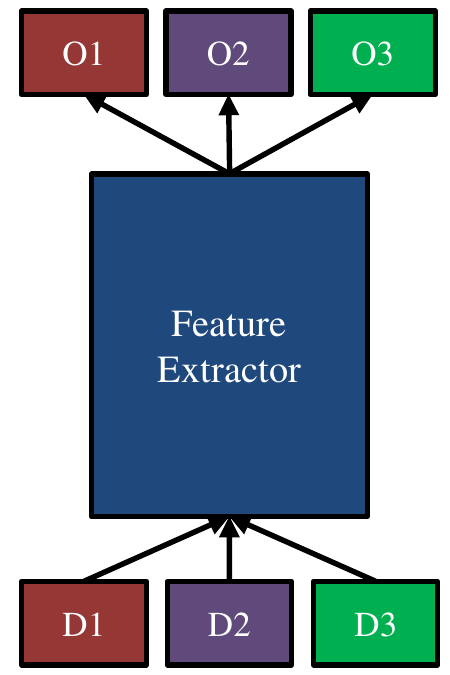}}{(c) Universal Detector}
\end{minipage}
\hfill
\begin{minipage}[t]{.18\linewidth}
\small
\centering
\centerline{\includegraphics[width=0.95\linewidth]{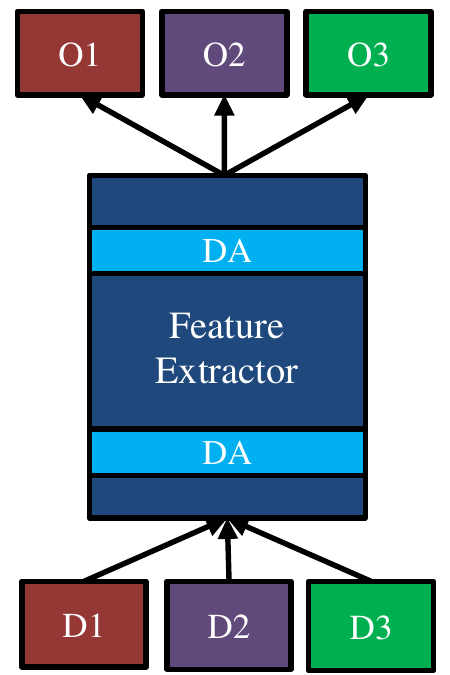}}{(d) Domain-attentive Universal Detector}
\end{minipage}
\vspace{2mm}
\scriptsize
\caption{Multi-domain and universal object detectors for three domains. ``D'' is the domain, ``O'' the output, ``A'' domain-specific adapter, and ``DA'' the proposed domain attention module. The blue color and the DA are domain-universal, but the other colors domain-specific.}
\label{fig:MDD}\vspace{-3mm}
\end{figure*}

There has been significant progress in object detection in recent years \cite{girshick2015fast,ren2015faster,cai2016unified,lin2017feature,he2017mask,cai18cascadercnn}, powered by the availability of challenging and diverse object detection datasets, e.g. PASCAL VOC \cite{everingham2015pascal}, COCO \cite{lin2014microsoft}, KITTI \cite{geiger2012we}, WiderFace \cite{yang2016wider}, etc. However, existing detectors are usually \textit{domain-specific}, e.g. trained and tested on a single dataset. This is partly due to the fact that object detection datasets are diverse and there is a nontrivial domain shift between them. As shown in Figure \ref{fig:demo},  detection tasks can vary in terms of categories (human face, horse, medical lesion, etc.), camera viewpoints (images taken from aircrafts, autonomous vehicles, etc.), image styles (comic, clipart, watercolor, medical), etc. In general, high detection performance requires a detector specialized on the target dataset.

This poses a significant problem for practical applications, which are not usually restricted to any one of the domains of Figure \ref{fig:demo}. Hence, there is a need for systems capable of detecting objects regardless of the domain in which images are collected. A simple solution is to design a \textit{specialized} detector for each domain of interest, e.g. use $D$ detectors trained on $D$ datasets, and load the detector specialized to the domain of interest at each point in time. This, however, may be impractical, for two reasons. First, in most applications involving autonomous systems the domain of interest can change frequently and is not necessarily known a priori. Second, the overall model size increases linearly with the number of domains $D$. A recent trend, known as general AI, is to request that a single universal model solves multiple tasks \cite{kaiser2017one,kokkinos2017ubernet,zamir2018taskonomy}, or the same task over multiple domains \cite{rebuffi2017learning,bilen2017universal}. However, existing efforts in this area mostly address image classification, rarely targeting the problem of object detection. The fact that modern object detectors are complex systems, composed of a backbone network, proposal generator, bounding box regressor, classifier, etc., makes the design of a universal object detector much more challenging than a universal image classifier.

In this work, we consider the design of an object detector capable of operating over multiple domains. We begin by establishing a new universal object detection benchmark, denoted as UODB, consisting of 11 diverse object detection datasets (see Figure \ref{fig:demo}). This is significantly more challenging than the Decathlon \cite{rebuffi2017learning}  benchmark for multi-domain recognition. To the best of our knowledge, we are the first to attack universal object detection using deep learning. We expect this new benchmark will encourage more efforts in the area. We then propose a number of architectures, shown in Figure \ref{fig:MDD}, to address the universal/multi-domain detection problem. 

The two architecture on the left of Figure \ref{fig:MDD} are multi-domain detectors, which require prior knowledge of the domain of interest. The two architectures on the right are universal detectors, with no need for such knowledge. When operating on an unknown domain, the multi-domain detector have to repeat the inference process with different sets of domain-specific parameters, while the universal detector performs inference only once. The detector of Figure \ref{fig:MDD} (a) is a bank of domain-specific detectors, with no sharing of parameters/computations. Multi-domain learning (MDL) \cite{joshi2012multi,Nam_2016_CVPR,kim2017learning,yang2014unified, jiang2008cross, dredze2010multi} improves on this, by sharing parameters across various domains, and adding small domain-specific layers. In \cite{rebuffi2017learning,bilen2017universal}, expensive convolutional layers are shared and complemented with light-weight domain-specific {\it adaptation layers\/}. 
Inspired by these, we propose a new class of light adapters for detection, based on the squeeze and excitation (SE) mechanism of~\cite{hu2017squeeze}, and denoted {\it SE adapters\/}. This leads to the {\it multi-domain detector\/} of Figure \ref{fig:MDD} (b), where domain-specific SE adapters are introduced throughout the network to compensate for domain shift. On UODB, this detector outperforms that of Figure \ref{fig:MDD} (a) with $\sim$5 times fewer parameters.

In contrast, the {\it universal detector\/} of Figure \ref{fig:MDD} (c) shares all parameters/computations (other than output layers) across domains. It consists of a \textit{single} network, which is always active. This is the most efficient solution in terms of parameter sharing, but it is difficult for a single model to cover many domains with nontrivial domain shifts. Hence, this solution underperforms the multi-domain detector of Figure \ref{fig:MDD} (b). To overcome this problem, we propose the {\it domain-attentive universal detector\/} of Figure~\ref{fig:MDD} (d). This leverages a novel domain attention (DA) module, in which a bank of the new universal SE adapters (active at all times) is first added, and a feature-based attention mechanism is then introduced to achieve domain sensitivity. This module learns to assign network activations to different domains, through the universal SE adapter bank, and soft-routs their responses by the domain-attention mechanism. This enables the adapters to specialize on individual domains. Since the process is data-driven, the number of domains does not have to match the number of datasets and datasets can span multiple domains. This allows the network to leverage shared knowledge across domains, which is not available in the common single-domain detectors. Our experiments, on the newly established UODB, show that this data-driven form of parameter/computation sharing enables substantially better multi-domain detection performance than the remaining architectures of Figure \ref{fig:MDD}.

\section{Related Work}

\noindent{\bf Object Detection:}
The two stage detection framework of the R-CNN \cite{girshick2014rich}, Fast R-CNN \cite{girshick2015fast} and Faster R-CNN \cite{ren2015faster} detectors has achieved great success in recent years. Many works have expanded this base architecture. For example, MS-CNN \cite{cai2016unified} and FPN \cite{lin2017feature} built a feature pyramid to effectively detect objects of various scales; the R-FCN \cite{dai2016r} proposed a position-sensitive pooling to achieve further speed-ups; and the Cascade R-CNN \cite{cai18cascadercnn} introduced a multi-stage cascade for high quality object detection. In parallel, single-stage object detectors, such as YOLO \cite{redmon2016you} and SSD \cite{liu2016ssd}, became popular for their fairly good performance and high speed. However, none of these detectors could reach high detection performance on more than one dataset/domain without finetuning. In the pre-deep learning era, \cite{khosla2012undoing} proposed a universal DPM \cite{felzenszwalb2010object} detector, by adding dataset specific biases to the DPM. But this solution is limited since DPM is not comparable to deep learning detectors. 

\noindent{\bf Multi-Task Learning:}
Multi-task learning (MTL) investigates how to jointly learn multiple tasks simultaneously, assuming a single input domain. Various multi-task networks \cite{kokkinos2017ubernet,zamir2018taskonomy,he2017mask,liu2017hierarchical,wang2015towards,zhang2017overview} have been proposed for joint solution of tasks such as object recognition, object detection, segmentation, edge detection, human pose, depth, action recognition, etc., by leveraging information sharing across tasks. However, the sharing is not always beneficial, sometimes hurting performance \cite{evgeniou2005learning,kato2008multi}. To address this, \cite{misra2016cross} proposed a cross-stitch unit, which combines tasks of different types, eliminating the need to search through several architectures on a per task basis.  \cite{zamir2018taskonomy}  studied the common structure and relationships of several different tasks.

\noindent{\bf Multi-Domain Learning/Adaptation:}
Multi-domain learning (MDL) addresses the learning of representations for multiple domains, known a priori \cite{joshi2012multi,nam2016learning}. It uses a combination of parameters that are shared across domains and domain-specific parameters. The latter are adaptation parameters, inspired by works on domain adaptation \cite{patel2015visual,long2015learning,rosenfeld2018incremental,mallya2018piggyback}, where a model learned from a source domain is adapted to a target domain. \cite{bilen2017universal} showed that multi-domain learning is feasible by simply adding domain-specific BN layers to an otherwise shared network. \cite{rebuffi2017learning} learned multiple visual domains with residual adapters, while \cite{rebuffi2018efficient} empirically studied efficient parameterizations. However, they build on BN layers and are not suitable for detection, due to the batch constraints of detector training. Instead, we propose an alternative SE adapters, inspired by ``Squeeze-and-Excitation'' ~\cite{hu2017squeeze},
to solve this problem.

\noindent{\bf Attention Module:}
\cite{vaswani2017attention} proposed a self-attention module for machine translation,
and similarly, \cite{wang2018non} proposed a non-local network for video classification, based on a spacetime dependency/attention mechanism. \cite{hu2017squeeze} focused on channel relationships, introducing the SE module to adaptatively recalibrate channel-wise feature responses, which achieved good results on ImageNet recognition. In this work, we introduce a domain attention module inspired by SE to make data-driven domain assignments of network activations, for the more challenging problem of universal object detection.

\section{Multi-domain Object Detection}
\label{sec:multi-domain}

The problem of multi-domain object detection is to detect objects on various domains.

\subsection{Universal Object Detection Benchmark}
\label{subsec:dataset}

To train and evaluate universal/multi-domain object detection systems, we established a new universal object detection benchmark (UODB) of $11$ datasets: Pascal VOC \cite{everingham2015pascal}, WiderFace \cite{yang2016wider}, KITTI \cite{geiger2012we}, LISA \cite{mogelmose2012vision}, DOTA \cite{xia2018dota}, COCO \cite{lin2014microsoft}, Watercolor \cite{inoue2018cross}, Clipart \cite{inoue2018cross}, Comic \cite{inoue2018cross}, Kitchen \cite{georgakis2016multiview} and DeepLesions \cite{yan2018deep}. This set includes the popular VOC \cite{everingham2015pascal} and COCO \cite{lin2014microsoft}, composed of images of everyday objects, e.g. bikes, humans, animals, etc. The 20 VOC categories are replicated on CrossDomain \cite{inoue2018cross} with three subsets of Watercolor, Clipart and Comic, with objects depicted in watercolor, clipart and comic styles, respectively. Kitchen \cite{georgakis2016multiview} consists of common kitchen objects, collected with an hand-held Kinect, while WiderFace \cite{yang2016wider} contains human faces, collected on the web. Both KITTI \cite{geiger2012we} and LISA \cite{mogelmose2012vision} depict traffic scenes, collected with cameras mounted on moving vehicles. KITTI covers the categories of vehicle, pedestrian and cyclist, while LISA is composed of traffic signs. DOTA \cite{xia2018dota} is a surveillance-style dataset, containing objects such as vehicles, planes, ships, harbors, etc. imaged from aerial cameras. Finally DeepLesion \cite{yan2018deep} is a dataset of lesions on medical CT images. A representative example of each dataset is shown in Figure \ref{fig:demo}. Some more details are summarized in Table \ref{table:datasets}. Altogether, UODB covers a wide range of variations in category, camera view, image style, etc, and thus establishes a good suite for the evaluation of universal/multi-domain object detection.

\subsection{Single-domain Detector Bank}

The Faster R-CNN \cite{ren2015faster} is used as the baseline architecture of all detectors proposed in this work. As a single-domain object detector, the Faster R-CNN is implemented in two stages. First, a region proposal network (RPN) produces preliminary class-agnostic detection hypotheses. The second stage processes these with a region-of-interest detection network to output the final detections.

\begin{figure}
  \includegraphics[width=\linewidth]{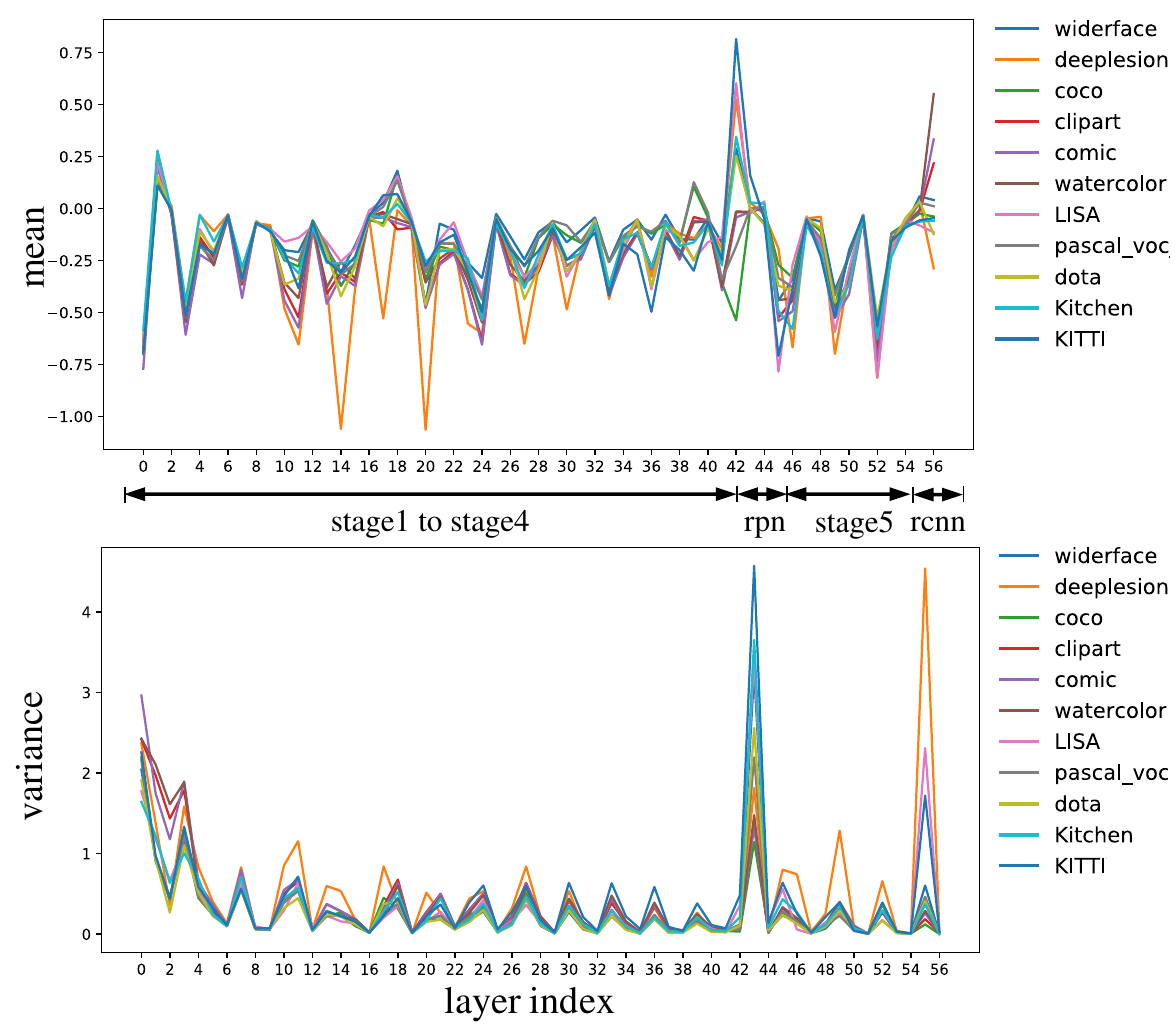}
  \caption{The activation statistics of all single-domain detectors.}
\label{fig:conv_statistical}\vspace{-3mm}
\end{figure}

As illustrated in Figure~\ref{fig:MDD} (a), the simplest solution to multi-domain detection is to use an independent detector per dataset. We use this detector bank as a multi-domain detection baseline. This solution is the most expensive, since it implies replicating all parameters of all detectors. Figure \ref{fig:conv_statistical} shows the statistics (mean and variance) of the convolutional activations of the $11$ detectors on the corresponding dataset. Some observations can be made. First, these statistics vary non-trivially across datasets. While the activation distributions of VOC and COCO are similar, DOTA, DeepLesion and CrossDomain have relatively different distributions. Second, the statistics vary across network layers. Early layers, which are more responsible for correcting domain shift, have more evident differences than latter layers. This tends to hold up to the output layers. These are responsible for the assignment of images to different categories and naturally differ. Interestingly, this behavior also holds for RPN layers, even though they are category-independent. Third, many layers have similar statistics across datasets. This is especially true for intermediate layers, suggesting that they can be shared by at least some domains.

\subsection{Adaptive Multi-domain Detector}
\label{subsec:semi-shared}

Inspired by Figure \ref{fig:conv_statistical}, we propose an adaptive multi-domain detector, shown in Figure \ref{fig:MDD} (b). In this model, the output and RPN layers are domain-specific. The remainder of the network, e.g. all convolutional layers, is shared. However, to allow adaptation to new domains, we introduce some additional domain-specific layers, as is commonly done in MDL~\cite{rebuffi2017learning,bilen2017universal}. These extra layers should be 1) sufficiently powerful to compensate for domain shift; 2) as light as possible to minimize  parameters/computation. The adaptation layers of \cite{rebuffi2017learning,bilen2017universal} rely extensively on BN. This is unfeasible for detection, where BN layers have to be frozen, due to the small batch sizes allowable for detector training. 

Instead, we have experimented with the squeeze-and-excitation (SE) module \cite{hu2017squeeze} of Figure \ref{fig:SE_and_multi_module} (a). There are a few reasons for this. First, feature-based attention is well known to be used in mammalian vision as a mechanism to adapt perception to different tasks and environments~\cite{yarbus1967eye, palmer1999vision, wolfe2000visual,itti2005principled,yantis1998control}. Hence, it seems natural to consider feature-based attention mechanisms for domain adaptation. Second, the SE is a module that accounts for interdependencies among channels to modulate channel responses. This can be seen as a a feature-based attention mechanism. Third the SE module has enabled the SENet to achieve state-of-the-art classification on ImageNet. Finally, it is a light-weight module. Even when added to each residual block of the ResNet~\cite{he2016deep} it increases the total parameter count by only $\sim$10\%. This is close to what was reported by \cite{rebuffi2017learning} for BN-based adapters. For all these reasons, we adopt the SE module as the atomic adaptation unit, used to build all domain adaptive detectors proposed in this work, and denote it by the {\it SE adapter\/}.

\subsection{SE Adapters}

\begin{figure}
\begin{minipage}[t]{.4\linewidth}
\centering
\centerline{\includegraphics[width=0.95\linewidth]{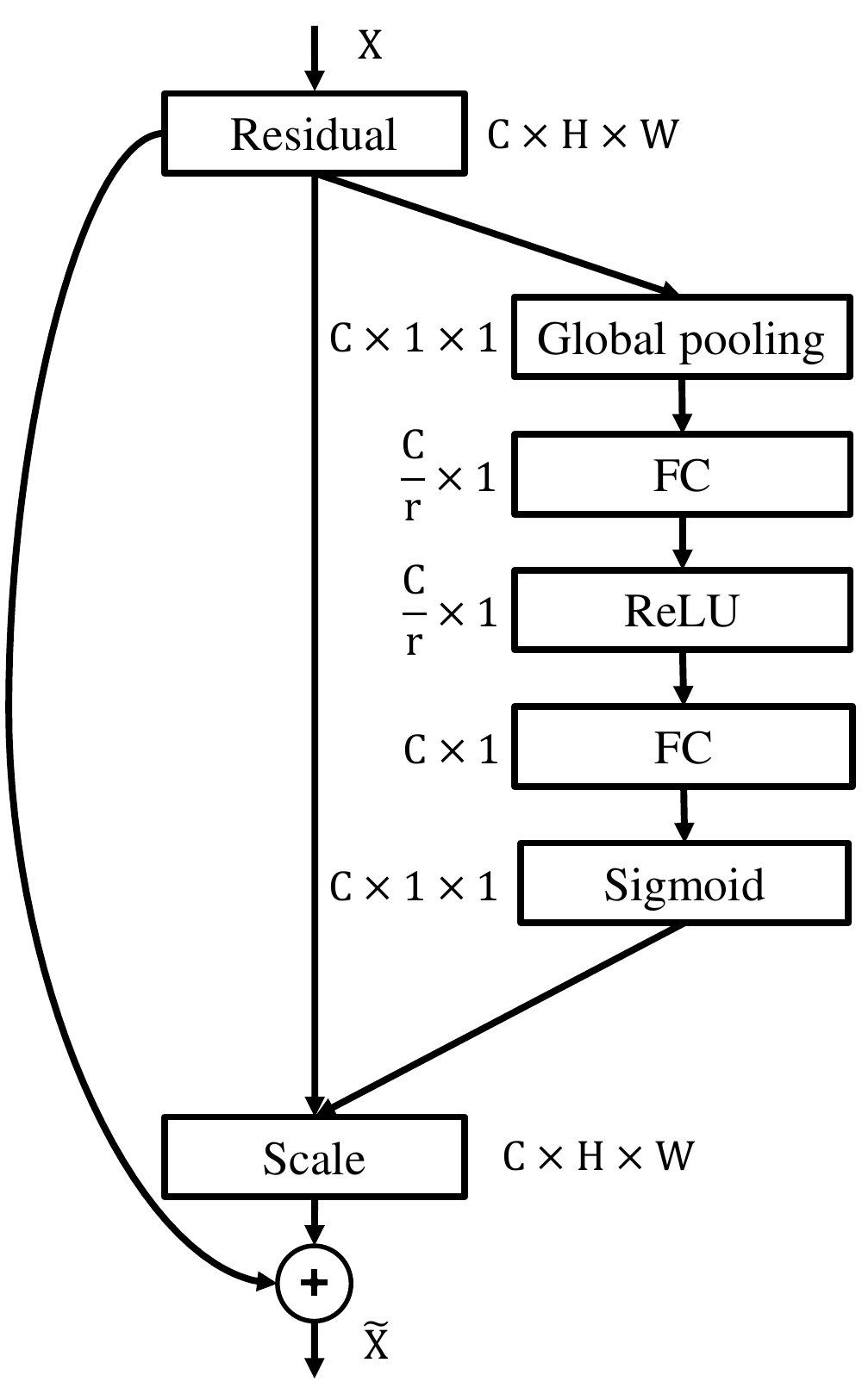}}{(a) SE adapter}
\end{minipage}
\hfill
\begin{minipage}[t]{.5\linewidth}
\centering
\centerline{\includegraphics[width=0.95\linewidth]{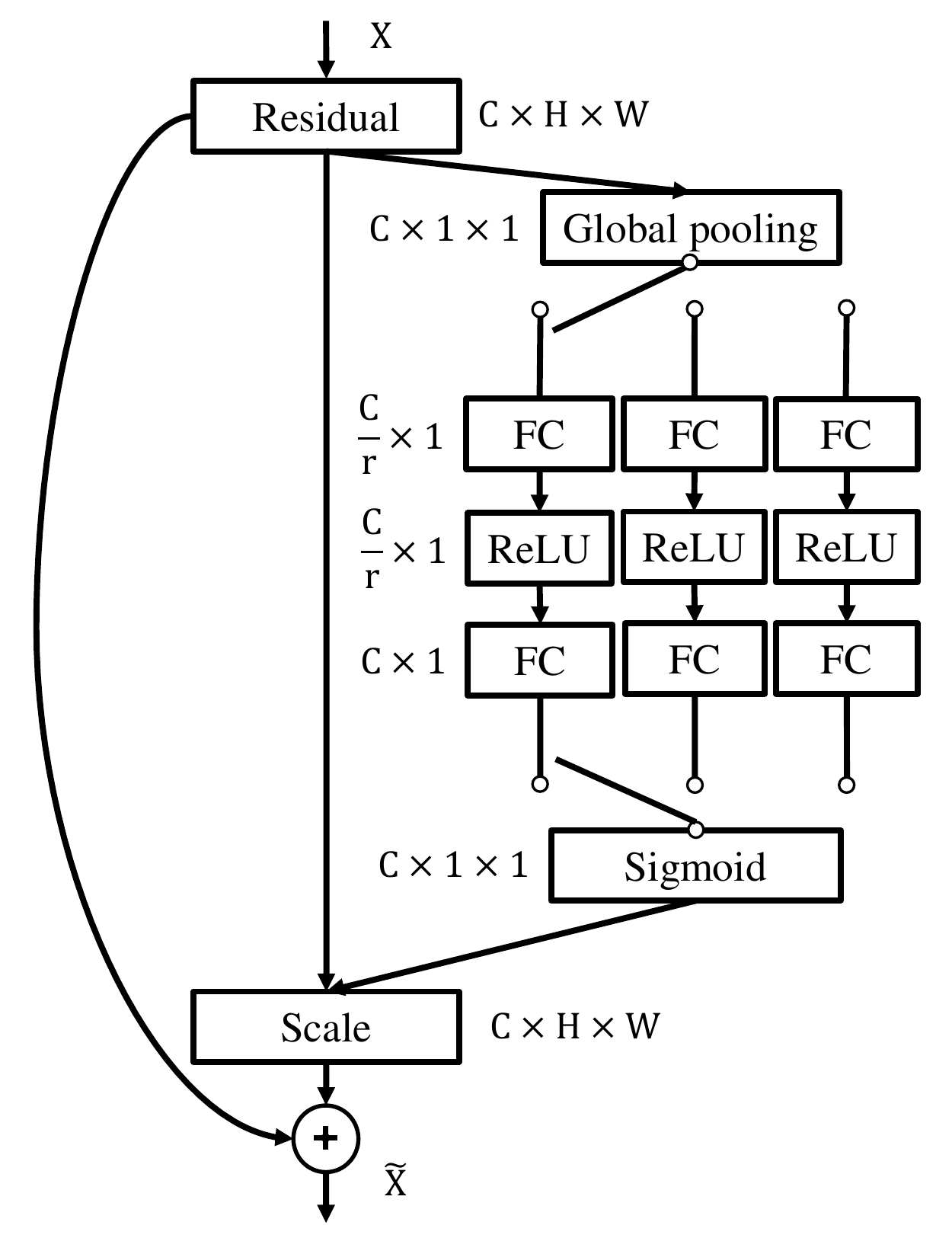}}{(b) SE adapter bank}
\end{minipage}
\vspace{2mm}
\caption{(a) block diagram of SE adapter and (b)  SE adapter bank.}
\label{fig:SE_and_multi_module}\vspace{-3mm}
\end{figure}

Following \cite{hu2017squeeze}, the {\it SE adapter\/} consists of the sequence of operations of Figure \ref{fig:SE_and_multi_module} (a): a global pooling layer, a fully connected (FC) layer, a ReLU layer, and a second FC layer, implementing the computation
\begin{equation}
\textbf{X}_{SE} = \textbf{F}_{SE}(\textbf{F}_{avg}(\textbf{X})), 
\label{eq:XSE}
\end{equation}
where $\textbf{F}_{avg}$ is a global average pooling operator, and $\textbf{F}_{SE}$ the combination of FC+ReLU+FC layers. The channel dimension reduction factor $r$, in Figure \ref{fig:SE_and_multi_module}, is set as 16 in our experiments. To enable multi-domain object detection, the SE adapter is generalized to the architecture of Figure \ref{fig:SE_and_multi_module} (b), which is denoted as the {\it SE adapter bank.\/} This consists of adding a SE adapter branch per domain and a domain-switch, which allows the selection of the SE adapter associated with the domain of interest. Note that this architecture assumes this domain to be known a priori. It leads to the  {\it multi-domain detector\/} of Figure \ref{fig:MDD} (b). Compared to Figure \ref{fig:MDD} (a), this model is up to 5 times smaller, while achieving better overall performance across the 11 datasets. 

\begin{figure*}
\begin{minipage}[t]{.69\linewidth}
\small
\centering
\centerline{\includegraphics[width=0.95\linewidth]{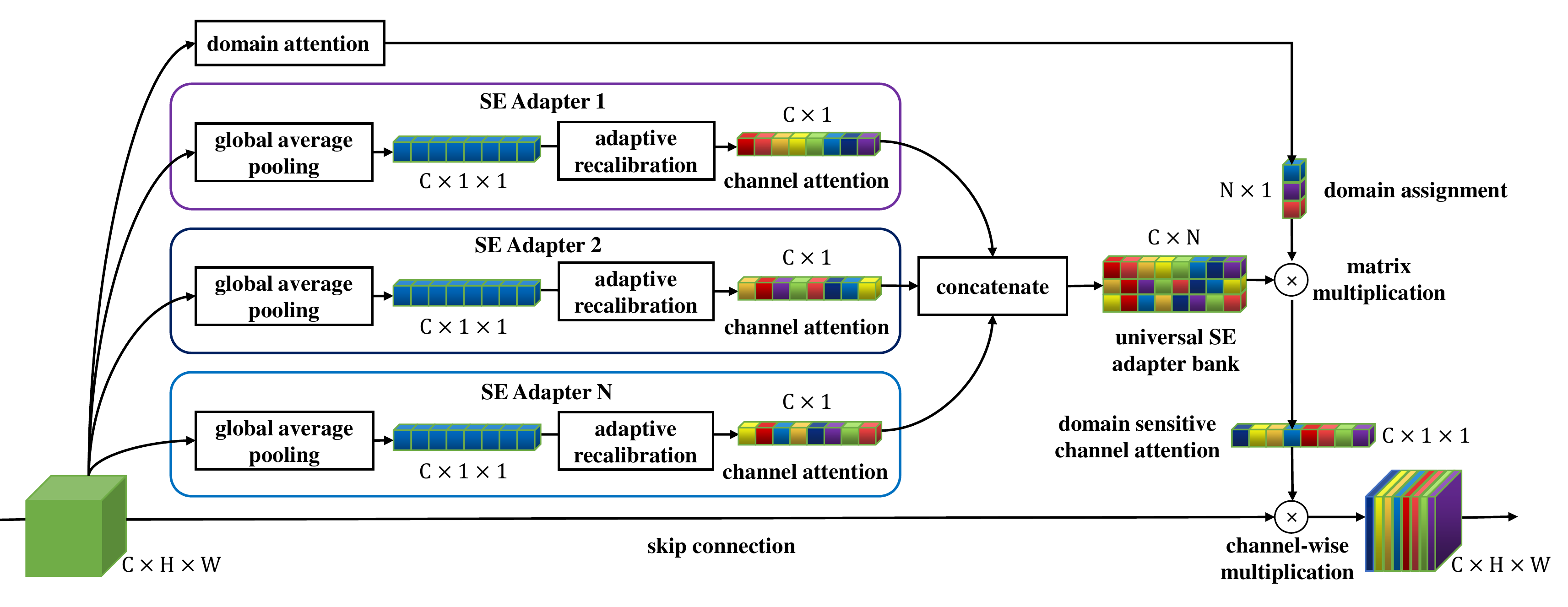}}
\end{minipage}
\hfill
\begin{minipage}[t]{.31\linewidth}
\small
\centering
\centerline{\includegraphics[width=0.95\linewidth]{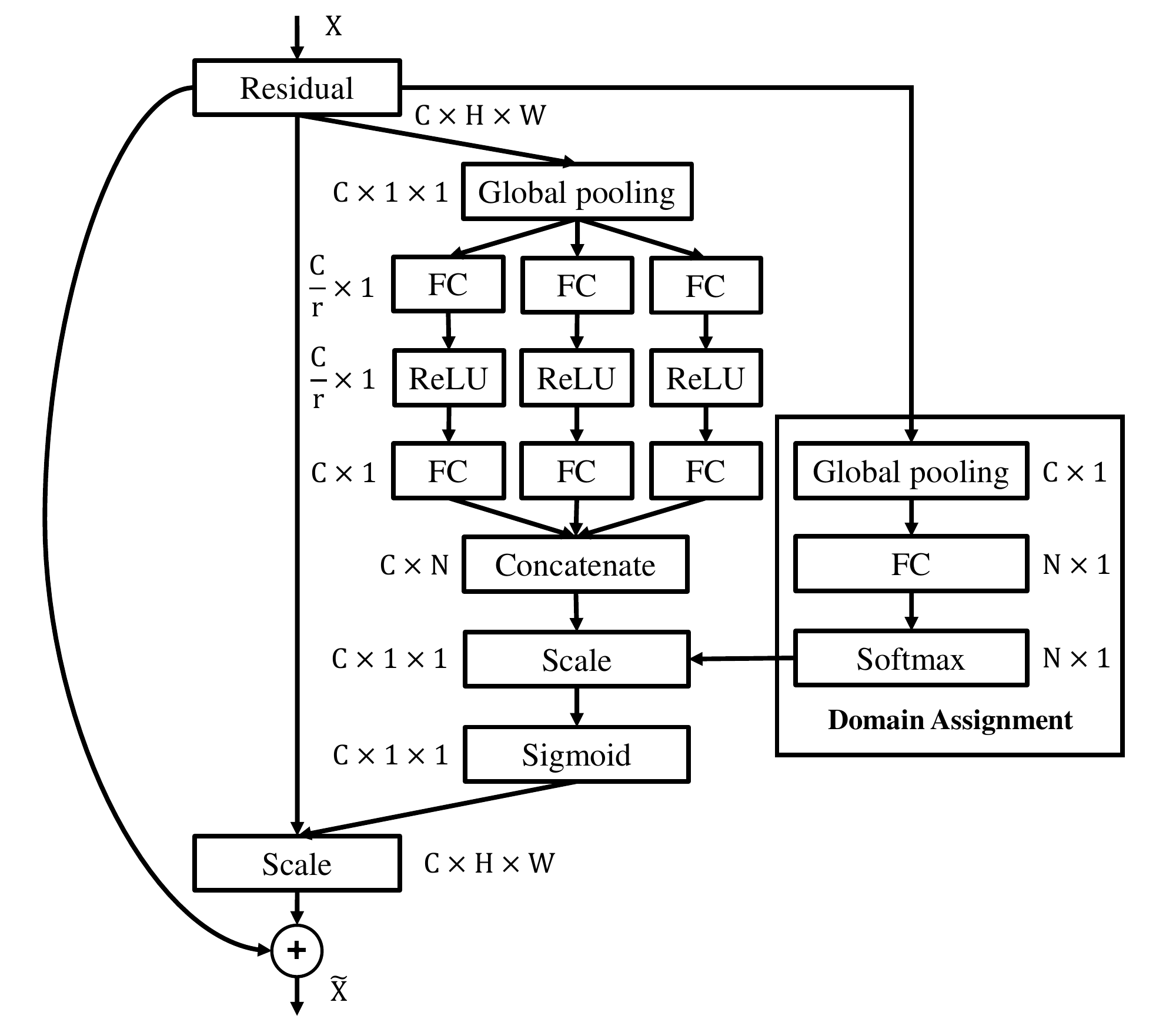}}
\end{minipage}
  \caption{The block diagram (left) and the detailed view (right) of the proposed domain adaptation module.}
\label{fig:CAM}\vspace{-3mm}
\end{figure*}

\section{Universal Object detection}

The detectors of the previous section require prior knowledge of the domain of interest. This is undesirable for autonomous systems, like robots or self-driving cars, where determining the domain is part of the problem to solve. In this section, we consider the design of {\it universal detectors,\/} which eliminate this problem.

\subsection{Universal Detector}

The simplest solution to universal detection, shown in Figure \ref{fig:MDD} (c), is to share a single detector by all tasks. Note that, even for this detector, the output layer has to be task-specific, by definition of the detection problem. This is not a problem because the task, namely what classes the system is trying to detect, is always known. Universality refers to the domain of input images that the detector processes, which does not have to be known in the case of Figure \ref{fig:MDD} (c). Beyond universal, the fully shared detector is the most efficient of all detectors considered in this work, as it has no domain-specific parameters. On the other hand, by forcing the same set of representations on all domains, it has little flexibility to deal with the statistical variations of Figure \ref{fig:conv_statistical}. In our experiments, this detector usually underperforms the multi-domain detectors of Figure \ref{fig:MDD} (a) and (b).

\vspace*{0.2cm}
\subsection{Domain-attentive Universal Detector}
\label{sec:CAM}

Ideally, a universal detector should have some domain sensitivity, and be able to adapt to different domains. We also have found that there is also a benefit in using task-specific RPN layers, due to the observations of Figure \ref{fig:conv_statistical}. While this has a lot in common with multi-domain detection, there are two main differences. First, the domain must be inferred automatically. Second, there is no need to tie domains and tasks. For example, the traffic tasks of Figure~\ref{fig:demo} operate on a common visual domain, ``traffic scenes'', which can have many sub-domains, e.g. due to weather conditions (sunny vs. rainy), environment (city vs. rural ), etc. Depending on the specific operating conditions, any of the tasks may have to be solved in any of the domains. In fact, the domains may not even have clear semantics, i.e. they can be data-driven. In this case, there is no need to request that each detector operates on a single domain, and a soft domain-assignment makes more sense. Given all of this, while domain adaptation can still be implemented with the SE adapter of Figure~\ref{fig:SE_and_multi_module} (a), the hard attention mechanism of Figure~\ref{fig:SE_and_multi_module} (b), which forces the network to fully attend to a single domain, can be suboptimal. To address this limitations, we propose the domain adaptation (DA) module of Figure \ref{fig:CAM}. This has two components, a {\it universal SE adapter bank\/} and a {\it domain attention \/} mechanism, which are discussed next. 

\subsection{Universal SE Adapter Bank}

The universal SE (USE) Adapter Bank, shown in Figure \ref{fig:CAM}, is an SE adapter bank similar to that of Figure~\ref{fig:SE_and_multi_module} (b). The main difference is that there is no domain switching, i.e. the adapter bank is {\it universal.\/} This is implemented by concatenating  the outputs of the individual domain adapters to form a universal representation space
\begin{equation}
\textbf{X}_{USE} = [\textbf{X}_{SE}^1,\textbf{X}_{SE}^2,...,\textbf{X}_{SE}^N] \in \mathbb{R}^{C \times N},
\end{equation}
where $N$ is the number of adapters and $\textbf{X}_{SE}^i$ the output of each adapter, given by~(\ref{eq:XSE}).
Note that $N$ is not necessarily identical to the number of detection tasks. The USE adapter bank can be seen as a non-linear generalization of the filter banks commonly used in signal processing \cite{vaidyanathan1993multirate}. Each branch (non-linearly) projects the input along a subspace matched to the statistics of a particular domain. The attention component then produces a domain-sensitive set of weights that are used to combine these projections in a data-driven way. In this case, there is no need to know the operating domain in advance. In fact there may not even be a single domain, since an input image can excite multiple SE adapter branches.

\subsection{Domain Attention}

The attention component, of Figure \ref{fig:CAM}, produces a domain-sensitive set of weights that are used to combine the SE bank projections. Motivated by the SE module, the domain attention component first applies a global pooling to the input feature map, to remove spatial dimensions, and then a softmax layer (linear layer plus softmax function)
\begin{equation} 
\textbf{S}_{DA} = \textbf{F}_{DA}(\textbf{X})=\text{softmax}(\textbf{W}_{DA}\textbf{F}_{avg}(\textbf{X})),
\end{equation}
where $\textbf{W}_{DA}\in \mathbb{R}^{N \times C}$ is the matrix of softmax layer weights. The vector $\textbf{S}_{DA}$ is then used to weigh the USE bank output $\textbf{X}_{USE}$, to produce
a vector of domain adaptive responses
\begin{equation}
\textbf{X}_{DA} = \textbf{X}_{USE}\textbf{S}_{DA}\in\mathbb{R}^{C \times 1}.
\end{equation}
As in the SE module of~\cite{hu2017squeeze}, $\textbf{X}_{DA}$ is finally used to channel-wise rescale the activations $\textbf{X} \in \mathbb{R}^{C \times H \times W}$ being adapted,
\begin{equation}
\widetilde{\textbf{X}} = \textbf{F}_{scale}(\textbf{X}, \sigma(\textbf{X}_{DA}))
\end{equation}
where \(\textbf{F}_{scale}(\cdot)\) implements a channel-wise multiplication, and $\sigma$ is the sigmoid function.

In this way, the USE bank captures the feature subspaces of the domains spanned by all datasets, and the DA mechanism soft-routes the USE projections. Both operations are data-driven, and operate with no prior knowledge of the domain. Unlike the hard attention mechanism of Figure \ref{fig:SE_and_multi_module} (b), this DA module enables information sharing across domains, leading to a more effective representation. In our experiments, the domain-attentive universal detector outperforms the other detectors of Figure \ref{fig:MDD}.

\begin{table}[t]
\tablestyle{2.0pt}{1.2}
\scriptsize
\begin{tabular}{l|x{18}x{32}x{28}|x{18}x{18}x{18}x{18}|x{20}}
\multirow{2}{*}{dataset} & \multicolumn{3}{c|}{dataset details} &\multicolumn{4}{c|}{hyperparameters}&\multirow{2}{*}{mAP}\\\cline{2-8}
&{class}  & {T/V/T} & domain & {size} & {$BS$} & {RoIs} & {$S/R$}\\ [.1em]
\shline
KITTI & 3 & 7k/-/7k & traffic & 576 & 256 & 128 & 12/3 & 64.3\\
WiderFace  & 1 & 13k/3k/16k & face & 800 & 256 & 256 & 12/1 & 48.9\\
VOC   & 20 & 8k/8k/5k & natural & 600 & 256 & 256 & 4/3 & 78.5\\
LISA  & 4 & 8k/-/2k & traffic & 800 & 64  & 32 & 4/3 & 88.3 \\
DOTA  & 15 & 14k/5k/10k & aerial & 600 & 128 & 128 & 12/3 & 57.5 \\
COCO  & 80 & 35k/5k/- & natural & 800 & 256 & 256 & 4/3 & 47.3\\
Watercolor  & 6 & 1k/-/1k & watercolor & 600 & 256 & 256 & 4/3 & 52.4\\
Clipart & 6 & 0.5k/-/0.5k & clipart & 600 & 256 & 256 & 4/3 & 32.1\\
Comic & 20 & 1k/-/1k & comic & 600 & 256 & 256 & 4/3 & 45.8\\
Kitchen & 11 & 5k/-/2k & indoor & 800 & 256 & 256 & 12/3 & 87.7\\
DeepLesion & 1 & 23k/5k/5k & medical & 512 & 128 & 64 & 12/3 & 51.3\\\hline
Average & - & - & - & - & - & - & - & 59.4\\
\end{tabular}\vspace{2mm}
\caption{The dataset details, the domain-specific hyperparameters and the performance of the single-domain detectors. ``T/V/T'' means train/val/test, ``size'' the shortest side of inputs, $BS$ RPN batch size, and $S/R$ anchor ``scales/aspect ratios''.}
\label{table:datasets}\vspace{-3mm}
\end{table}

\section{Experiments}
\label{sec:exp}

In all experiments, we used a PyTorch implementation \cite{jjfaster2rcnn} of the Faster R-CNN with the SE-ResNet-50\cite{hu2017squeeze}/ResNet50\cite{he2016deep} pretrained on ImageNet, as the backbone for all detectors. Training started with a learning rate of 0.01 for 10 epochs and 0.001 for another 2 epochs on 8 synchronized GPUs, each holding 2 images per iteration. All samples of a batch are from a single (randomly sampled) dataset, and in each epoch, all samples of each dataset are processed only once. As is common for detection, the first convolutional layer, the first residual block and all BN layers are frozen, during training. These settings were used in all experiments, unless otherwise noted. Both multi-domain and universal detectors were trained on all domains of interest simultaneously.

The Faster R-CNN has many hyperparameters. In the literature, where detectors are tested on a single domain, these are tuned to the target dataset, for best performance. This is difficult, and very tedious, to do over the $11$ datasets now considered. We use the same hyperparameters across datasets, except when this is critical for performance and relatively easy to do, e.g. the choice of anchors. The main dataset-specific hyperparameters are shown in Table \ref{table:datasets}.

\subsection{Datasets and Evaluation}

Our experiments used the new UODB benchmark introduced in Section \ref{subsec:dataset}. For Watercolor \cite{inoue2018cross}, Clipart \cite{inoue2018cross}, Comic \cite{inoue2018cross}, Kitchen \cite{georgakis2016multiview} and DeepLesion \cite{yan2018deep}, we trained on the official \texttt{trainval} sets and tested on the \texttt{test} set. For Pascal VOC \cite{everingham2015pascal}, we trained on VOC2007 and VOC2012 \texttt{trainval} set and tested on VOC2007 \texttt{test} set. For WiderFace \cite{yang2016wider}, we trained on the \texttt{train} set and tested on the \texttt{val} set. For KITTI \cite{geiger2012we}, we followed the train/val splitting of \cite{cai2016unified} for development and trained on the \texttt{trainval} set for the final results on \texttt{test} set. For LISA \cite{mogelmose2012vision}, we trained on the \texttt{train} set and tested on the \texttt{val} set. For DOTA \cite{xia2018dota}, we followed the pre-processing of \cite{xia2018dota}, trained on \texttt{train} set and tested on \texttt{val} set. For MS-COCO \cite{lin2014microsoft}, we trained on COCO 2014 \texttt{valminusminival} and tested on \texttt{minival}, to shorten the experimental period.

All detectors were evaluated on each dataset individually. The Pascal VOC mean average precision (mAP) was used for evaluation in all cases. The average mAPs was used as the overall measure of universal/multi-domain detection performance. The domain attentive universal detector was also evaluated using the official evaluation tool of each dataset, for comparison with the literature.

\subsection{Single-domain Detection}

Table \ref{table:datasets} shows the results of the single-domain detector bank of Figure \ref{fig:MDD} (a) on all datasets. Our VOC baseline with the SE-ResNet-50 is 78.5, and better than the Faster R-CNN performance of \cite{ren2017faster,he2016deep} (76.4 mAP for ResNet-101). The other entries in the table are incomparable to the literature, where different evaluation metrics/tools are used for different datasets. The detector bank is a fairly strong baseline for multi-domain detection (average mAP of 59.4). 

\begin{table}[t]
\tablestyle{2.0pt}{1.2}
\scriptsize
\begin{tabular}{l|x{32}|x{16}|x{18}x{15}x{28}x{18}x{21}|x{17}}
& Params & time & KITTI & VOC & WiderFace & LISA & Kitchen & Avg\\ [.1em]
\shline
single-domain & 155.1M & 5x & 64.3 & 78.5 & 48.8 & 88.3 & 87.7 & 73.5\\
adaptive & 42.37M & 6x & 67.8  & 78.9 &  49.9  & \bd{88.5} & 86.0 & 74.2\\
BNA \cite{bilen2017universal} &31.72M  & 5x & 64.0 & 71.9 & 44.0 &  66.8 & 84.3 & 66.2\\
RA \cite{rebuffi2017learning} & 82.72M  & 6x  & 64.3 & 70.5 & 46.9 &  69.1 & 84.6 & 67.1\\
universal & 29.58M & 1x & 64.2  & 75.0 & 43.5 & 88.9 & 86.8 & 71.5\\
universal+DA$\dag$ & 37.47M & 1.3x  & 67.5 & 79.0 & 49.8 & 88.2 & 88.0 & 74.6\\
universal+DA& 37.56M & 1.33x  & \bd{67.9} & \bd{79.2} & \bd{52.2} & 87.5 & \bd{88.5} & \bd{75.1}\\
\end{tabular}\vspace{2mm}
\caption{The comparison on multi-domain detection. $\dag$ denotes fixed assignment. ``time'' is the relatively run-times on the five datasets when the domain is unknown.} 
\label{table:multi_domain}\vspace{-5mm}
\end{table}

\begin{table*}[t]
\tablestyle{2.0pt}{1.2}
\scriptsize
\begin{tabular}{l|x{30}|x{36}|x{36}|x{20}x{20}x{28}x{20}x{24}x{20}x{20}x{32}x{20}x{20}x{31}|x{26}}
& {\# adapters} & {Params} & {DA index} & {KITTI} & {VOC} & {WiderFace} & {LISA} & {Kitchen} &{COCO} & {DOTA} & {DeepLesion} & {Comic} & {Clipart} & {Watercolor} & {Avg}\\ [.1em]
\shline
single-domain  & -      & 340.7M      & -    & 64.3  & 78.5   & 48.8  & 88.3 & 87.7 & \bf{47.3}  & \bf{57.5} & 51.2  & 45.8 & 32.1 & 52.6 & 59.4\\
adaptive  & 11      & 58.13M      & -    & 68.0  & 82.1   & 50.6  & \bf{88.5} & 87.2 & 45.7  & 54.1 & 53.0  & 50.0 & 56.1 & 57.8 & 63.0\\
universal & -      & 32.60M      & -       & 67.5  & 80.9  & 45.5   & 87.1   & 88.5  & 45.5 &  54.7 & 45.3    & 51.1    & 43.1    &  47.0   &  59.7 \\
universal+DA  & 11     & 58.29M      & all    & \bf{68.1}   & 82.0   & 51.6  &  88.3 & \bf{90.1} & 46.5 & 57.0 & \bf{57.3}  & 50.7 & 53.1 & 58.4 & 63.8\\
universal+DA*  & 6      & 41.74M      & first+middle        & 67.6      & \bf{82.7}    & \bf{51.8}    &  87.9  & 88.7 & 46.8 & 57.0 & 54.8 &  52.6  &  54.6   &  58.2   &  63.9 \\
universal+DA*  & 8      & 44.80M      & first+middle        & 68.0      & 82.4    & 51.3    &  87.6  & 90.0 & 47.0 & 56.3 & 53.4 &  \bf{53.4}  &  \bf{55.8}   &  \bf{60.6}   &  \bf{64.2} \\
\end{tabular}\vspace{2mm}
\caption{Overall results on the full universal object detection benchmark (11 datasets).}
\label{table:11_datasets}\vspace{-5mm}
\end{table*}

\begin{table}[t]
\tablestyle{2.0pt}{1.2}
\scriptsize
\begin{tabular}{x{30}|x{32}|x{22}x{22}x{28}x{22}x{22}|x{22}}
\# adapters & Params & KITTI & VOC & WiderFace & LISA & Kitchen & Avg\\ [.1em]
\shline
single & 155.1M & 64.3 & 78.5 & 48.8 & 88.3 & 87.7 & 73.5\\\hline
3 &34.82M  & 67.6 & 78.0 & 51.9 & 88.1 & 87.1 & 74.5\\
5 &37.88M  & \bd{67.9} & 79.2 & \bd{52.2} & 87.5 & 88.5 & 75.1\\
7 &40.94M  & \bd{67.9} & \bd{79.6} & \bd{52.2} & \bd{89.5} & \bd{88.7} & \bd{75.6}\\
9 &44.01M  & 67.7 & 79.2 & \bd{52.2} & \bd{89.5} & 86.9 & 75.1\\
\end{tabular}\vspace{2mm}
\caption{The effect of SE adapters number.}
\label{table:adapter_num}\vspace{-3mm}
\end{table}

\subsection{Multi-domain Detection}

Table \ref{table:multi_domain} compares the multi-domain object detection performance of all architectures of Figure \ref{fig:MDD}. For simplicity, only five datasets (VOC, KITTI, WiderFace, LISA and Kitchen) were used in this section. The table confirms that the adaptive multi-domain detector of Section \ref{subsec:semi-shared} (``adaptive'') is light-weight, only adding $\sim$11M parameters to the Faster R-CNN over the five datasets. Nevertheless, it outperforms the much more expensive single-domain detector bank by 0.7 points. Note that the latter is a strong baseline, showing the multi-domain detector can beat individually trained models with a fraction of the computation. Table \ref{table:multi_domain} also shows that the proposed SE adapter significantly outperforms the BN adapter (BNA) of \cite{bilen2017universal}  and the residual adapter (RA) or \cite{rebuffi2017learning}, previously proposed for classification. This is not surprising, given the above discussed inadequacy of BN as an adaptation mechanism for object detection. 

The universal detector of Figure \ref{fig:MDD} (c) is even more efficient, adding only 0.5M parameters to the Faster R-CNN, accounting for domain-specific RPN and output layers. However, its performance (``universal'' in Table \ref{table:multi_domain}) is much weaker than that of the adaptive multi-domain detector (1.7 points). Finally, the domain-attentive universal detector (``universal+DA'') has the best performance. With a $\sim$7\% parameter increase  per domain, i.e. comparable to the multi-domain detector, it outperforms the single-domain bank baseline by 1.6 points. To assess the importance of data-driven domain attention mechanism of Figure \ref{fig:CAM} (b), we fixed the soft domain assignments, simply averaging the SE adapter responses, during both training and inference. This (denoted ``universal+DA$\dag$'') caused a performance drop of 0.5 point. Finally, Table \ref{table:multi_domain} shows the relative run-times of all methods on the five datasets, when the domain is unknown. It can be seen that ``universal+DA'' is about 4$\times$ faster than the multi-domain detectors (``single-domain'' and ``adaptive'') and only 1.33$\times$ slower than ``universal''.

\subsection{Effect of the number of SE adapters}

For the USE bank of Figure \ref{fig:CAM} (b), the number $N$ of SE adapters does not have to match the number of detection tasks. Table \ref{table:adapter_num} summarizes how the performance of the domain attentive universal detector depends on $N$. For simplicity, we again use  5 datasets in this experiment. As we can see in Table \ref{table:adapter_num}, more SE adapters will not always lead to better performance. Generally, performance improves with the number of adapters until we reached 7 adapters, adding 9 adapters will only get similar performance with 5 adapters and increase parameters number a lot. On the other hand, the number of parameters increases linearly with the number of adapters. In these experiments, the best trade-off between performance and parameters is around 5 adapters. This suggests that, while a good rule of thumb is to use ``as many adapters as domains'', fewer adapters can be used when complexity is at a premium. 

\subsection{Beneficial/harmful relations among 3 domains}
Since the underlying properties of the datasets are unknown, it is difficult to know which datasets will benefit each other and which will not. For example, VOC and COCO are close to each other, and it is widely known COCO will benefit VOC. But what about VOC and DeepLesion? Probably they will hurt each other, given the nontrivial source difference between them (web image v.s. medical CT image). Some control experiments are shown in Table \ref{table:beneficial_3_datasets} to explore the beneficial/harmful relations among domains, in which KITTI and VOC were always used with a third dataset to control. As expected, DeepLesion has the least benefits to improve KITTI and VOC, while it is interesting to find LISA has the most benefits for KITTI and VOC.

\begin{table}[t]
\tablestyle{2.0pt}{1.2}
\scriptsize
\begin{tabular}{x{45}|x{32}x{22}x{22}x{28}x{22}x{22}}
& KITTI & VOC & LISA & Lesion & Kitchen & DOTA \\ [.1em]
\shline
single-domain & 64.3 & 78.5 & 88.3 & 51.2 & 87.7 & 57.5 \\\hline
\multirow{4}{*}{universal+DA}  & 69.2 & 79.8 & 89.2 & - &  - & -\\
                      & 67.4 & 78.2 & - & 55.5 &  - & -\\
                      & 67.4 & 78.7 & - & - & 85.2 & -\\
                      & 67.8 & 78.5 & - & - & - & 54.8\\
\end{tabular}\vspace{2mm}
\caption{The beneficial/harmful relationships among 3 domains.}
\label{table:beneficial_3_datasets}\vspace{-3mm}
\end{table}

\begin{figure}
\begin{center}
  \includegraphics[width=0.48\textwidth]{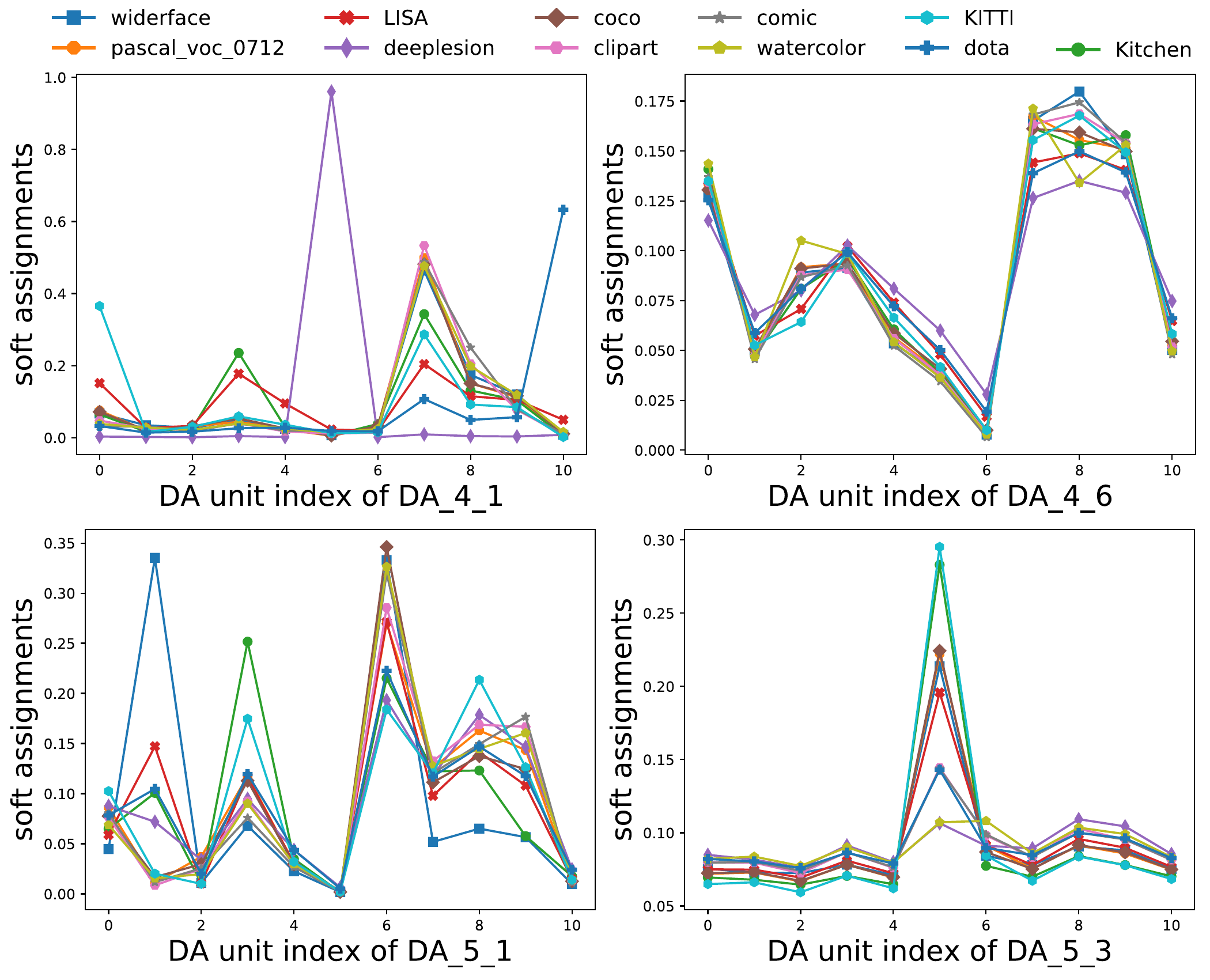}
  \caption{Soft assignments across SE units for all datasets.}
\label{fig:came_weight}
\end{center}\vspace{-7mm}
\end{figure}

\begin{table*}[t]\vspace{-3mm}
\subfloat[The comparison on VOC 2007 \texttt{test}. $\ddagger$/$\dagger$ denotes with COCO \texttt{trainval}/\texttt{val}.\label{table:pascal}]{
\tablestyle{2pt}{1.05}\scriptsize \begin{tabular}{l|x{44}|x{20}}
 & {Backbone} & {mAP}\\
\shline
Faster-RCNN \cite{ren2015faster} & ResNet-101 & 76.4 \\
R-FCN \cite{dai2016r} & ResNet-50 &77.0 \\
Faster-RCNN$\ddagger$ \cite{ren2017faster} & VGG16 &  78.8 \\
\hline
Faster-RCNN (ours) & ResNet-50 & 78.0 \\
Faster-RCNN (ours) & SE-ResNet-50 & 78.5 \\
Faster-RCNN+DA & DA-ResNet-50 &79.6\\
Faster-RCNN+DA$\dagger$ & DA-ResNet-50 &82.7\\
\end{tabular}}\hspace{1mm}
\subfloat[The comparison on WiderFace \texttt{Val}.\label{table:widerface}]{\tablestyle{4.2pt}{1.05}\scriptsize\begin{tabular}{l|x{44}|x{20}x{20}x{20}}
  & { Backbone} & {Easy}& { Medium}& {Hard}\\
\shline
Faster-RCNN \cite{ren2015faster} & VGG-16 &  0.907 & 0.850 & 0.492\\
MS-CNN \cite{cai2016unified} & VGG-16 &  0.916 &  0.903 &  0.802 \\
HR \cite{hu2017finding} & ResNet-101 &  0.925 &  0.910 &  0.806 \\
SSH \cite{najibi2017ssh} & VGG-16 &  0.931 &  0.921 &  0.845 \\
\hline
Faster-RCNN (ours) & ResNet-50 &  0.905 & 0.864 & 0.548 \\
Faster-RCNN (ours) & SE-ResNet-50 &  0.910 & 0.872 & 0.556 \\
Faster-RCNN+DA & DA-ResNet-50 &  0.914 & 0.882 & 0.587\\
\end{tabular}}\hspace{1mm}
\subfloat[Sensitivity at 4 FPs per image on DeepLesion \texttt{test} set. \label{table:DeepLesion}]{
\tablestyle{2pt}{1.05}\scriptsize\begin{tabular}{l|x{44}|x{26}}
 & {Backbone} & {Sensitivity}\\
\shline
Faster-RCNN \cite{ren2015faster} & VGG-16 & 81.62 \\
R-FCN \cite{dai2016r} & VGG-16 & 82.21 \\
3-DCE, 9 Slices \cite{yan20183d} & VGG-16 & 84.34 \\
3-DCE, 27 Slices \cite{yan20183d} & VGG-16 & 85.65 \\
\hline
Faster-RCNN (ours) & ResNet-50 &  81.34 \\
Faster-RCNN (ours) & SE-ResNet-50 &  82.44 \\ 
Faster-RCNN+DA & DA-ResNet-50 &  87.29\\
\end{tabular}}\vspace{-2mm}\\
\subfloat[The comparison on Clipart, Watercolor and Comic \texttt{test} set.\label{table:cross_domain}]{
\tablestyle{4pt}{1.05}\scriptsize\begin{tabular}{l|x{44}|x{26}x{26}x{26}}
 & {Backbone} & {Clipart} & {Watercolor} & {Comic}\\
\shline
ADDA \cite{tzeng2017adversarial} & VGG-16 &  27.4 & 49.8 & 49.8\\
Faster-RCNN\cite{ren2015faster} & VGG-16 &  26.2 & - & -\\
SSD300 \cite{liu2016ssd} & VGG-16 &  26.8 & 49.6 & 24.9\\
Faster-RCNN+DT+PL\cite{inoue2018cross} & VGG-16 &  34.9 & - & -\\
SSD300+DT+PL\cite{inoue2018cross} & VGG-16 &  46.0 & 54.3 & 37.2\\
\hline
Faster-RCNN (ours) & ResNet-50 & 36.7 & 51.2 & 42.6 \\
Faster-RCNN (ours) & SE-ResNet-50 & 32.1 & 52.6 & 45.8\\
Faster-RCNN+DA & DA-ResNet-50 & 54.6 & 58.2 & 52.6\\
\end{tabular}}\hspace{10mm}
\subfloat[The comparison on KITTI \texttt{test} set of car\label{table:kitti}.]{
\tablestyle{4pt}{1.05}\scriptsize\begin{tabular}{l|x{44}|x{26}x{26}x{26}}
 & { Backbone} & { Moderate}& { Easy}& { Hard}\\
\shline
Faster-RCNN \cite{ren2015faster} & VGG-16 &  81.84 & 86.71 & 71.12\\
SDP+CRC \cite{yang2016exploit} & VGG-16 &  83.53  &  90.33 &  71.13 \\
YOLOv3 \cite{redmon2018yolov3} & Darknet-53 &  84.13 &  84.30&  76.34\\
MS-CNN \cite{cai2016unified} & VGG-16 &  88.83 &  90.46 &  74.76 \\
F-PointNet \cite{qi2018frustum} & PointNet &  90.00 &  90.78 &  80.80\\
\hline
Faster-RCNN (ours) & ResNet-50 &  80.28 & 90.43 & 70.9 \\
Faster-RCNN (ours) & SE-ResNet-50 &  81.83 & 90.34 & 71.23\\
Faster-RCNN+DA & DA-ResNet-50 &  88.23 & 90.45 & 74.21\\
\end{tabular}}\vspace{2mm}
\caption{The comparison with official evaluation on Pascal VOC, KITTI, DeepLesion, Clipart, Watercorlor, Comic and WiderFace.}
\label{table:offcial_results}\vspace{-3mm}
\end{table*}

\subsection{Results on the full benchmark}

Table~\ref{table:11_datasets} presents results on the full benchmark. The settings are as above, but we used 10 epochs with learning rate 0.1, and then 4 epochs with 0.01 on 8 GPUs, each holding 2 images. For universal detector without our proposed module, training with learning rate 0.1, even with warming up for 1 epoch, will always get loss exploding, therefore, we will use 0.1\(\times\)lr, i.e. 0.01, to train universal detector. The universal detector performs comparably to the single-domain detector bank, with $10$ times fewer parameters. The domain-attentive universal detector (``universal+DA'') improves baseline performance by 4.4 points with a $5$-fold parameter decrease. It has large performance gains ($>$5 points) on DeepLesion, Comic, and Clipart. This is because Comic/Clipart contain underpopulated classes, greatly benefiting from information leveraged from other domains. The large gain of DeepLesion is quite interesting, given the nontrivial domain shift between its medical CT images and the RGB images of the other datasets. The gains are mild for VOC, KITTI, Kitchen, WiderFace and WaterColor (1$\sim$5 points), and none for COCO, LISA and DOTA. In contrast, for the universal detector, joint training is not always beneficial. This shows the importance of domain sensitivity for universal detection. 

To investigate what was learned by the domain attention module of Figure \ref{fig:CAM} (b), we show the soft assignments of each dataset, averaged over its validation set, in Figure \ref{fig:came_weight}. Only the first and last blocks of the 4th and 5th residual stages are shown. The fact that some datasets, e.g. VOC and COCO, have very similar assignment distributions, suggests a substantial domain overlap. On the other hand, DOTA and DeepLesion have distributions quite distinct from the remaining. For example, on block ``DA\_4\_1'', DeepLesion fully occupies a single domain. These observations are consistent with Figure \ref{fig:conv_statistical}, indicating that the proposed DA module is able to learn domain-specific knowledge.

A comparison of the first and the last blocks of each residual stage, e.g. ``DA\_4\_1'' v.s. ``DA\_4\_6'', shows that the latter are much less domain sensitive than the former, suggesting that they could be made universal. To test this hypothesis, we trained a model with only 6 SE adapters for the 11 datasets, and only in the first and middle blocks, e.g. ``DA\_4\_1'' and ``DA\_4\_3'', all the other blocks will only add 1 SE adapter without domain assignment module. This model, ``universal+DA*'', achieved the best performance with much less parameters than the ``universal+DA'' detector of 11 adapters. It outperformed the single domain baseline by 4.5 points. We also trained a model with 8 SE adapters, adding another two adapters will get another 0.3 points increase, outperforms single-domain baseline by 4.8 points.

\subsection{Official evaluation}

Since, to the best of our knowledge, this is the first work to explore universal/multi-domain object detection on 11 datasets, there is no literature for a direct comparison. Instead, we compared the ``universal+DA*'' detector of Table \ref{table:11_datasets} to the literature  using the official evaluation for each dataset. This is an unfair comparison, since the universal detector has to remember 11 tasks. All single-domain detector will use SE-ResNet50 as backbone.
On VOC, we trained two models, with/without COCO. Results are shown in Table \ref{table:pascal}, where all methods were trained on Pascal VOC 07+12 \texttt{trainval}. Note that our Faster R-CNN baseline (SE-ResNet-50 backbone) is stronger than that of \cite{he2016deep} (ResNet-101). Adding universal domain adapters improved on the baseline by more than 1.1 points.
Adding COCO enabled another 3.1 points. Note that, 1) this universal training is different from the training scheme of \cite{ren2017faster} (the network trained on COCO then finetuned on VOC), where the final model is only optimized for VOC; and 2) only the 35k images of COCO2014  \texttt{valminusminival} were used. 

The baseline was the default Faster R-CNN that initially worked on VOC, with minimum dataset-specific changes, e.g. in Table \ref{table:datasets}. Table \ref{table:kitti} shows that this performed weakly on KITTI. However, the addition of adapters, enabled a gain of 6.4 points (\texttt{Moderate} setting). This is comparable to detectors optimized explicitly on KITTI, e.g. MS-CNN \cite{cai2016unified} and F-PointNet \cite{qi2018frustum}. For WiderFace, which has enough training face instances, the gains of shared knowledge are smaller (see Table \ref{table:widerface}). On the other hand, on DeepLesion and CrossDomain (Clipart, Comic and Watercolor), see Table \ref{table:DeepLesion} and \ref{table:cross_domain} respectively, the domain attentive universal detector significantly outperformed the state-of-the-art. Although DeepLesion contains more than 33k images, most of them only contain one lesion, which will make detector unable to get adequate training. For Cross-domain datasets, they only contain a few thousands samples. Single-domain detector is easy to be over-fitting when fine-tuned on so small datasets. Jointly training with other datasets, especially with VOC and MS-COCO which share some categories with cross-domain datasets, will benefit the representation learning process a lot for Cross-domain datasets and mitigate over-fitting problem. Overall, these results show that a single detector, which operates on 11 datasets, is competitive with single-domain detectors in highly researched datasets, such as VOC or KITTI, and substantially better than the state-of-the-art in less explored domains. This is achieved with a relatively minor increase in parameters, vastly smaller than that needed to deploy 11 single task detectors. 

\section{Conclusion}

We have investigated the unexplored and challenging problem of universal/multi-domain object detection. We proposed a universal detector that requires no prior domain knowledge, consisting of a \textit{single} network that is active for all tasks. The proposed detector achieves domain sensitivity through a novel data-driven domain adaptation module and was shown to outperform multiple universal/multi-domain detectors on a newly established benchmark, and even individual detectors optimized for a single task.

\paragraph{Acknowledgment} This work was partially funded by NSF awards IIS-1546305 and IIS-1637941, a gift from 12 Sigma Technologies, and NVIDIA GPU donations.


{\small
\bibliographystyle{ieee_fullname}
\bibliography{egbib}
}

\end{document}